\pdfoutput=1
\documentclass[11pt,a4paper]{article}
\usepackage[hyperref]{acl2020}
\usepackage{times}
\usepackage{latexsym}
\usepackage{linguex}
\usepackage{amsmath}
\usepackage{amssymb}
\usepackage{booktabs}
\usepackage{linguex}
\usepackage{url}
\usepackage{array}
\newcolumntype{x}[1]{>{\centering\arraybackslash\hspace{0pt}}p{#1}}
\newcolumntype{C}[1]{>{\centering\let\newline\\\arraybackslash\hspace{0pt}}m{#1}}
\usepackage{tabularx}
\usepackage{graphicx}
\usepackage{dblfloatfix}
\usepackage{enumitem}
\usepackage{ textcomp }
\usepackage{subcaption}

\usepackage{microtype}
\usepackage{booktabs}

\aclfinalcopy 
\def\aclpaperid{1856} 

\title{How Can We Accelerate Progress Towards Human-like Linguistic Generalization?}

\author{Tal Linzen\\
  Department of Cognitive Science\\
  Johns Hopkins University\\
  \texttt{tal.linzen@jhu.edu}}

\date{}

\begin{document}
\maketitle
\begin{abstract}
This position paper describes and critiques the Pretraining-Agnostic Identically Distributed (PAID) evaluation paradigm, which has become a central tool for measuring progress in natural language understanding. This paradigm consists of three stages: (1)~pre-training of a word prediction model on a corpus of arbitrary size; (2)~fine-tuning (transfer learning) on a training set representing a classification task; (3)~evaluation on a test set drawn from the same distribution as that training set. This paradigm favors simple, low-bias architectures, which, first, can be scaled to process vast amounts of data, and second, can capture the fine-grained statistical properties of a particular data set, regardless of whether those properties are likely to generalize to examples of the task outside the data set. This contrasts with humans, who learn language from several orders of magnitude less data than the systems favored by this evaluation paradigm, and generalize to new tasks in a consistent way. We advocate for supplementing or replacing PAID with paradigms that reward architectures that generalize as quickly and robustly as humans. 
\end{abstract}


\section{Introduction}

The special session of the 2020 Annual Meeting of Association for Computational Linguistics invites us to take stock of the progress made in the field in the last few years. There is no question that we have made significant progress in a range of applications: current machine translation systems for high-resource languages, for example, are undeniably better than those we had a decade ago. This opinion piece will focus on a different question: are we making progress towards the classic goal of mimicking human linguistic abilities in machines---towards a model that acquires language as efficiently as humans, and generalizes it as humans do to new structures and contexts (``tasks'')?

I will argue that an evaluation paradigm that has rapidly established itself as one of the main tools for measuring progress in the field---a paradigm I will term, for want of a catchier name, Pretraining-Agnostic Identically Distributed evaluation (PAID)---encourages progress in a direction that is at best orthogonal to the goal of human-like generalization. Because it does not consider sample efficiency, this approach rewards models that can be trained on massive amounts of data, several orders of magnitude more than a human can expect to be exposed to. And because benchmark scores are computed on test sets drawn from the same distribution as their respective training sets, this paradigm favors models that excel in capturing the statistical patterns of particular data sets over models that generalize as a human would.

\section{Human-like Generalization}
\label{sec:human}

Humans learn language from much more limited exposure than most contemporary NLP systems. An analysis of recordings taken in the environment of the child of an MIT professor between the ages of~9~and~24~months found that the child heard or produced approximately eight million words over this 15-month period \cite{roy2015predicting}. Children in lower socioeconomic status families in Western societies receive significantly less linguistic input than that (around 3 million words per year, \citealt{hart1995meaningful}); even more strikingly, members of the Tsimane community in Bolivia spend about 15~times less time per hour speaking to their children than do highly educated American families \cite{cristia2019child}. If NLP systems were as sample-efficient as Tsimane children, far fewer languages would be considered ``low-resource languages''.

Despite the limited amount of exposure to their language, humans generalize their linguistic knowledge in a consistent way to structures that are infrequent or non-existent in corpora  \cite{sprouse2013comparison}, and quickly learn to do new things with language (what we sometimes refer to in NLP as ``tasks''). As I discuss below, this is not the case for current deep learning systems: when tested on cases sampled from a distribution that differs from the one they were trained on, their behavior is unpredictable and inconsistent with that of humans \cite{jia2017adversarial,mccoy2019right}, and they require extensive instruction on each new task \cite{yogatama2019learning}. Humans' rapid and consistent generalization abilities rely on powerful inductive biases, which likely arise from a combination of innate building blocks and experience with diverse learning problems \cite{lake2017building}.

Systems that generalize like humans would be useful not only for NLP, but also for the scientific study of human language acquisition and processing \cite{keller2010cognitively,dupoux2018cognitive}. But, as I will argue in the next two sections, it is unclear whether our dominant evaluation paradigms are getting us closer to this goal.

\section{Pretraining-Agnostic Evaluation}

Over the last two years, deep learning systems have obtained rapidly increasing scores on language understanding benchmarks such as GLUE \cite{wang2019glue} or SuperGLUE \cite{superglue}. These benchmarks aggregate multiple supervised classification tasks---such as sentiment analysis, linguistic acceptability judgments, or entailment detection---and collate the scores obtained on those tasks into a leaderboard, with a single headline score for each model averaging its scores on each individual task. For each of these classification tasks, a data set that was generated by a particular process, often involving crowdsourcing, is randomly split into two: a training set, which the system is allowed to observe, and a held-out test set, on which it is evaluated.

A standard recipe has emerged for achieving high scores on such benchmarks. A neural network---typically, one based on the  transformer architecture \cite{vaswani2017attention}---is pretrained on a denoising objective, such as filling in one or more blanks in a vast number of sentences. This network is then fine-tuned (performs transfer learning) on the benchmark's supervised tasks, each of which include a much smaller number of training examples than the pretraining corpus \cite{howard2018universal,peters2018elmo}. The T5 model \cite{raffel2019exploring}---the system that boasted the highest score on SuperGLUE at the time of writing---achieved an average accuracy of 88.9\% on this benchmark, slightly lower than that of untrained human annotators (89.8\%), and more than 20 percentage points higher than the score obtained just a few months earlier by BERT \cite{devlin2019bert,superglue}. This jump in accuracy does not reflect significant modeling innovations: both BERT and T5 are transformers trained on similar objectives that differ primarily in their scale.

When ranking systems, leaderboards such as SuperGLUE do not take into account the amount of pretraining data provided to each model. Pretraining corpora are not standardized, and the amount of pretraining data is not always easy to discern from the papers reporting on such systems. Here is my attempt to reconstruct the recent evolution of pretraining corpus sizes.\footnote{Corpus sizes reported in massive-corpus pretraining papers are often specified in gigabytes, or number of model-specific subword units, instead of measures such as number of words that are easier to compare across articles. My estimates are based on an average English word length of 4.7~characters and a space or punctuation mark after each word.} BERT, uploaded to arXiv in October 2018, was trained on 3.3 billion words; \mbox{XLNet} (\citeauthor{yang2019xlnet}, June 2019), was trained on 78 GB of text, or approximately 13 billion words; RoBERTa (\citeauthor{liu2019roberta}, July 2019) was trained on 160~GB of text, or around 28~billion words; and T5 (\citeauthor{raffel2019exploring}, October 2019) was trained on  750~GB of text, or approximately 130~billion words.

When we rely on a single leaderboard to compare systems trained on corpora with such a large range of sizes, we are not comparing architectures, but rather interactions of architectures, corpus sizes, and computational resources available for training. While this may be a useful comparison for an engineer who seeks to plug an existing trained model into a larger pipeline, this approach is unlikely to advance us towards the goal advocated in this article. The 130 billion word corpus that T5 was trained on is much larger than the corpus that a human can expect to be exposed to before adulthood (fewer than 100~million words, see Section~\ref{sec:human}). But a leaderboard that evaluates only bottom-line transfer learning accuracy inherently disadvantages a sample-efficient model pretrained on a few dozen million words compared to a model such as T5. For all we know, it is possible that architectures rewarded by PAID, such as massive transformers, \textit{only} work well when given an amount of data that is orders of magnitude greater than that available to humans. If that is the case, our exploration of the space of possible models could be going in a direction that is orthogonal to the one that might lead us to models that can imitate humans' sample efficiency (one example of such direction is neural networks with explicit symbolic structure, which are harder to scale up, but perform well on smaller data sets: \citealt{kuncoro2018lstms,wilcox2019structural}).

\section{Identically Distributed Training Set and Test Set}

The remaining two letters of the PAID acronym refer to the practice of evaluating success on classification tasks using training and test set generated using the same process. Typically, a single data set is collected and is randomly split into a training portion and test portion. While this may seem reasonable from a machine learning perspective, it has become clear that this form of evaluation obscures possible mismatches between the generalizations that we as humans believe a system performing the task should acquire, and the generalizations that the system in fact extracts from the data.

Consider, for example, crowdsourced natural language inference (NLI) data sets, in which workers are asked to generate a sentence that contradicts the prompt shown to them \cite{bowman2015large}. One strategy that crowdworkers adopt when generating a contradiction is to simply negate the prompt, for example by inserting the word \textit{not}. This strategy is often effective: \textit{the man is sleeping} contradicts \textit{the man is not sleeping}. Conversely, it is much less likely that the worker would use the word \textit{not} when asked to generate a sentence that is entailed by the prompt. Taken together, such worker choices lead to a strong correlation between the presence of the word \textit{not} in the hypothesis and the label \textsc{contradiction}. It would be surprising if low-bias learners such as neural networks did not notice such a correlation, and indeed they do, leading them to respond \textsc{contradiction} with high probability any time the hypothesis contains a negation word \cite{gururangan2018annotation,poliak2018hypothesis}. Of course, relying on the presence of the word \textit{not} is not a generally valid inference strategy; for example, \textit{the man is awake} entails, rather than contradicts, \textit{the man is not sleeping}. 

Numerous generalization issues of this sort have been documented, for NLI and for other tasks. In the syntactic domain, \citet{mccoy2019right} showed that BERT fine-tuned on the crowdsourced MultiNLI data set \cite{williams2018multinli} achieves high accuracy on the MultiNLI test set, but shows very little sensitivity to word order when tested on constructed examples that require an analysis of the structure of the sentence; for example, this model is likely to conclude that \textit{the detective followed the suspect} entails \textit{the suspect followed the detective}. 

In short, the models, unable to discern the intentions of the data set's designers, happily recapitulate any statistical patterns they find in the training data. With a random training/test split, any correlation observed in the training set will hold approximately for the test set, and a system that learned it could achieve high test set accuracy. And indeed, we have models that excel in the PAID paradigm, even exceeding the performance of human annotators on the test portion of the corpus used for fine-tuning \cite{nangia2019bowman}, but, when tested on controlled examples, make mistakes that a human would rarely make.\footnote{Comparisons between human annotators and transformers are arguably unfair: before observing the test set, the models receive hundreds of thousands of examples of the output of the data-generating process. This contrasts with humans annotators, who need to perform the task based on their general language understanding skills. It would be an entertaining though somewhat cruel experiment to repeat the comparison after matching the amount of exposure that humans and pretrained transformers receive to the quirks of the data set.}

The generalizations that a statistical model extracts from the data are always the result of the interaction between the model's inductive biases and the statistical properties of the data set. In the case of BERT's insensitivity to word order in NLI, the model does not seem to have a strong inductive bias one way or another; its sensitivity to word order varies widely depending on the weight initialization of the fine-tuning classifier and the order of the fine-tuning examples \cite{mccoy2019berts}, and its syntactic behavior in the inference task can be made to be more consistent with human intuitions if the training set is augmented to include a larger number of examples illustrating the importance of word order \cite{min2020augmentation}. While BERT is capable of learning to use syntax for inference given a sufficiently strong signal, then, it prefers to use other heuristics, if possible. This contrasts with human-like generalization in this task, which would likely start from the assumption that any language understanding task should recruit our knowledge of syntax: it would most likely be difficult to convince humans to \textit{ignore} syntax when understanding a sentence, as BERT does.

\section{The Generalization Leaderboard}

What is the way forward? My goal is not to argue that there is no value to the leaderboard approach, where a single number or a small set of numbers can be used to quickly compare models. Despite the drawbacks of this approach---in particular, its tendency to obscure the fine-grained strengths and weaknesses of particular models, as I discuss below---hill climbing on a metric can enable a productive division of labor between groups that develop strong benchmarks, groups that propose new models and inference methods, and groups that have the engineering skills and computational resources necessary to train those models on the number of GPUs they require to thrive.

Instead, my argument is that the current division of labor is unproductive. At the risk of belaboring the mountaineering metaphor, one might say that groups with access to engineering and computing resources are climbing the PAID hill, while other groups, which document the same models' unreliable generalization behavior---or retrain them on smaller data sets to produce the learning curves that are often missing from engineering papers---are climbing the interpretability track hill, producing papers that are more and more sophisticated and well-respected but do not influence the trajectory of mainstream model development. This section describes some design decisions that can lead to better alignment between the two sets of research groups. Many of these points are not new---in fact, some of these properties were standard in evaluation paradigms 10 or 20 years ago---but are worth revisiting given recent evaluation trends.

\paragraph{Standard, moderately sized pretraining corpora.} To complement current evaluation approaches, we should develop standard metrics that promote sample efficiency. At a minimum, we should standardize the pretraining corpus across all models, as some CoNLL shared tasks do. Multiple leaderboards can be created that will measure performance on increasingly small subsets of this pretraining corpus size---including ones that are smaller than 100 million words. To make stronger contact with the human language acquisition literature, a leaderboard could compare models on their ability to learn various linguistic generalizations from the CHILDES repository of child-directed speech \cite{macwhinney2000tools}.

\paragraph{Independent evaluation in multiple languages.} A model can be sample-efficient for English, but not for other languages. We should ensure that our architectures, like humans learners, are not optimized for English \cite{bender2011achieving}. To do so, we should develop matched training corpora and benchmarks for multiple languages. A composite score could reflect average performance across languages \cite{hu2020xtreme}. In keeping with our goal of mimicking humans, who are known for their ability to learn any language without learning English first, we should train and test the models separately on each language, instead of focusing on transfer from English to other languages---an important, but distinct, research direction.

\paragraph{What about grounding?} In response to studies comparing training corpus sizes between deep learning models and humans (e.g., \citealt{vanschijndel2019quantity}), it is sometimes pointed out that humans do not  learn language from text alone---we also observe the world and interact with it. This, according to this argument, renders the comparison meaningless. While the observation that children learn from diverse sources of information is certainly correct, it is unclear whether any plausible amount of non-linguistic input could offset the difference between 50 million words (humans) and 130 billion words (T5). Instead of taking this observation as a \textit{carte blanche} to ignore sample efficiency, then, we should address it experimentally, by collecting multimodal data sets \cite{suhr2019nlvr2,hudson2018gqa}, developing models that learn from them efficiently, and using the Generalization Leaderboard to measure how effective this signal is in aligning the model's generalization behavior with that of humans.

\paragraph{Normative evaluation.} Performance metrics should be derived not from samples from the same distribution as the fine-tuning set, but from what we might term normative evaluation: expert-created controlled data sets that capture our intuitions about how an agent should perform the task \cite{marelli2014sick,marvinlinzen18,warstadt2019blimp,ettinger2020bert}. Such data sets should be designed to be difficult to solve using heuristics that ignore linguistic principles.  While experts are more expensive than crowdworkers, the payoff in terms of data set quality is likely to be considerable. In parallel, we should continue to explore approaches such as adversarial filtering that may limit crowdworkers' ability to resort to shortcuts \cite{zellers2018swag,nie2019adversarial}.

Normative evaluation is related to but distinct from adversarial evaluation. Adversarial attacks usually focus on a specific trained model, starting from an example that the model classifies correctly, and perturbing it in ways that, under the normative definition of the task, should not affect the classifier's decision. For example, adversarial evaluation for a given question answering system may take an existing instance from the data set, and find an irrelevant sentence that, when added to the paragraph  that the question is about, changes the system's response \cite{jia2017adversarial}.  By contrast, the goal of the normative evaluation paradigm is not to fool a particular system by exploiting its weaknesses, but simply to describe the desirable performance on the task in a unambiguous way.

\paragraph{Test-only benchmarks.} A central point that bears repeating is that we should not fine-tune our models on the evaluation benchmark. Despite our best efforts, we may never be able to create a benchmark that does not have unintended statistical regularities. Fine-tuning on the benchmark may clue the model into such unintended correlations \cite{liu2019inoculation}. Any pretrained model will still need to be taught how to perform the transfer task, of course, but this should be done using a separate data set, perhaps one of those that are currently aggregated in GLUE. Either way, the Generalization Leaderboard should favor models that, like humans, are able to perform tasks with minimal instruction (few-shot learning, \citealt{yogatama2019learning}).

\paragraph{What about efficiency?} The PAID paradigm is agnostic not only to pretraining resources, but also to properties of the model such as the number of parameters, the speed of inference, or the number of GPU hours required to train it. These implementational-level factors \cite{marr1982vision} are orthogonal to our generalization concerns, which are formulated at the level of input--output correspondence. If efficiency is a concern, however, such properties can be optimized directly by modifying pretraining-agnostic benchmarks to take them into account \cite{schwartz2019green}.

\paragraph{Breakdown by task and phenomenon.} Benchmarks should always provide a detailed breakdown of accuracy by task and linguistic phenomenon: a model that obtains mediocre average performance, but captures a particular phenomenon very well, can be of considerable interest. Discouragingly, even though GLUE reports such task-specific scores---and even includes diagnostic examples created by experts---these finer-grain results have failed to gain the same traction as the headline GLUE benchmark. Other than exhorting authors to pay greater attention to error analysis in particular and linguistics in general---granted, an exhortation without which no ACL position piece can be considered truly complete---we should insist, when reviewing papers, that authors include a complete breakdown by phenomenon as an appendix, and discuss noteworthy patterns in the results. For authors that strongly prefer that their paper include a headline number that is larger than numbers reported in previous work, the leaderboard could offer alternative headline metrics that would reward large gains in one category even when those are offset by small losses in others.

\section{Conclusion}

I have described the currently popular Pretraining-Agnostic Identically Distributed paradigm, which selects for models that can be  trained easily on an unlimited amount of data, and that excel in capturing arbitrary statistical patterns in a fine-tuning data set. While such models have considerable value in applications, I have advocated for a parallel evaluation ecosystem---complete with a leaderboard, if one will motivate progress---that will reward models for their ability to generalize in a human-like way. Human-like inductive biases will improve our models' ability to learn language structure and new tasks from limited data, and will align the models' generalization behavior more closely with human expectations, reducing the allure of superficial heuristics that do not follow linguistic structure, and the prevalence of adversarial examples, where changes to the input that are insignificant from a human perspective turn out to affect the network's behavior in an undesirable way.

\bibliography{acl_position}

\begin{thebibliography}{43}
\expandafter\ifx\csname natexlab\endcsname\relax\def\natexlab#1{#1}\fi

\bibitem[{Bender(2011)}]{bender2011achieving}
Emily~M. Bender. 2011.
\newblock \href
  {http://journals.linguisticsociety.org/elanguage/lilt/article/view/2624.html}
  {On achieving and evaluating language-independence in {NLP}}.
\newblock \emph{Linguistic Issues in Language Technology}, 6(3):1--26.

\bibitem[{Bowman et~al.(2015)Bowman, Angeli, Potts, and
  Manning}]{bowman2015large}
Samuel~R. Bowman, Gabor Angeli, Christopher Potts, and Christopher~D. Manning.
  2015.
\newblock \href {https://doi.org/10.18653/v1/D15-1075} {A large annotated
  corpus for learning natural language inference}.
\newblock In \emph{Proceedings of the 2015 Conference on Empirical Methods in
  Natural Language Processing}, pages 632--642, Lisbon, Portugal. Association
  for Computational Linguistics.

\bibitem[{Cristia et~al.(2019)Cristia, Dupoux, Gurven, and
  Stieglitz}]{cristia2019child}
Alejandrina Cristia, Emmanuel Dupoux, Michael Gurven, and Jonathan Stieglitz.
  2019.
\newblock \href {https://doi.org/10.1111/cdev.12974} {Child-directed speech is
  infrequent in a forager-farmer population: a time allocation study}.
\newblock \emph{Child Development}, 90(3):759--773.

\bibitem[{Devlin et~al.(2019)Devlin, Chang, Lee, and
  Toutanova}]{devlin2019bert}
Jacob Devlin, Ming-Wei Chang, Kenton Lee, and Kristina Toutanova. 2019.
\newblock \href {https://www.aclweb.org/anthology/N19-1423/} {{BERT}:
  Pre-training of deep bidirectional transformers for language understanding}.
\newblock In \emph{Proceedings of the 2019 Conference of the North {A}merican
  Chapter of the Association for Computational Linguistics: Human Language
  Technologies, Volume 1 (Long and Short Papers)}, pages 4171--4186,
  Minneapolis, Minnesota. Association for Computational Linguistics.

\bibitem[{Dupoux(2018)}]{dupoux2018cognitive}
Emmanuel Dupoux. 2018.
\newblock \href {https://doi.org/10.1016/j.cognition.2017.11.008} {Cognitive
  science in the era of artificial intelligence: A roadmap for
  reverse-engineering the infant language-learner}.
\newblock \emph{Cognition}, 173:43--59.

\bibitem[{Ettinger(2020)}]{ettinger2020bert}
Allyson Ettinger. 2020.
\newblock \href {https://doi.org/10.1162/tacl_a_00298} {What {BERT} is not:
  Lessons from a new suite of psycholinguistic diagnostics for language
  models}.
\newblock \emph{Transactions of the Association for Computational Linguistics},
  8:34--48.

\bibitem[{Gururangan et~al.(2018)Gururangan, Swayamdipta, Levy, Schwartz,
  Bowman, and Smith}]{gururangan2018annotation}
Suchin Gururangan, Swabha Swayamdipta, Omer Levy, Roy Schwartz, Samuel Bowman,
  and Noah~A. Smith. 2018.
\newblock \href {https://www.aclweb.org/anthology/N18-2017/} {Annotation
  artifacts in natural language inference data}.
\newblock In \emph{{Proceedings of the 2018 Conference of the North American
  Chapter of the Association for Computational Linguistics: Human Language
  Technologies, Volume 2 (Short Papers)}}, pages 107--112. Association for
  Computational Linguistics.

\bibitem[{Hart and Risley(1995)}]{hart1995meaningful}
Betty Hart and Todd~R. Risley. 1995.
\newblock \emph{Meaningful differences in the everyday experience of young
  American children.}
\newblock Baltimore: P. H. Brookes.

\bibitem[{Howard and Ruder(2018)}]{howard2018universal}
Jeremy Howard and Sebastian Ruder. 2018.
\newblock \href {https://doi.org/10.18653/v1/P18-1031} {Universal language
  model fine-tuning for text classification}.
\newblock In \emph{Proceedings of the 56th Annual Meeting of the Association
  for Computational Linguistics (Volume 1: Long Papers)}, pages 328--339,
  Melbourne, Australia. Association for Computational Linguistics.

\bibitem[{Hu et~al.(2020)Hu, Ruder, Siddhant, Neubig, Firat, and
  Johnson}]{hu2020xtreme}
Junjie Hu, Sebastian Ruder, Aditya Siddhant, Graham Neubig, Orhan Firat, and
  Melvin Johnson. 2020.
\newblock \href {https://arxiv.org/abs/2003.11080} {Xtreme: A massively
  multilingual multi-task benchmark for evaluating cross-lingual
  generalization}.
\newblock arXiv preprint 2003.11080.

\bibitem[{Hudson and Manning(2019)}]{hudson2018gqa}
Drew~A. Hudson and Christopher~D. Manning. 2019.
\newblock \href {https://arxiv.org/abs/1902.09506} {{GQA}: A new dataset for
  real-world visual reasoning and compositional question answering}.
\newblock \emph{Conference on Computer Vision and Pattern Recognition (CVPR)}.

\bibitem[{Jia and Liang(2017)}]{jia2017adversarial}
Robin Jia and Percy Liang. 2017.
\newblock \href {http://aclweb.org/anthology/D17-1215} {Adversarial examples
  for evaluating reading comprehension systems}.
\newblock In \emph{{Proceedings of the 2017 Conference on Empirical Methods in
  Natural Language Processing}}, pages 2021--2031. Association for
  Computational Linguistics.

\bibitem[{Keller(2010)}]{keller2010cognitively}
Frank Keller. 2010.
\newblock \href {https://www.aclweb.org/anthology/P10-2012} {Cognitively
  plausible models of human language processing}.
\newblock In \emph{Proceedings of the {ACL} 2010 Conference Short Papers},
  pages 60--67, Uppsala, Sweden. Association for Computational Linguistics.

\bibitem[{Kuncoro et~al.(2018)Kuncoro, Dyer, Hale, Yogatama, Clark, and
  Blunsom}]{kuncoro2018lstms}
Adhiguna Kuncoro, Chris Dyer, John Hale, Dani Yogatama, Stephen Clark, and Phil
  Blunsom. 2018.
\newblock \href {https://www.aclweb.org/anthology/P18-1132/} {{LSTMs} can learn
  syntax-sensitive dependencies well, but modeling structure makes them
  better}.
\newblock In \emph{{Proceedings of the 56th Annual Meeting of the Association
  for Computational Linguistics (Volume 1: Long Papers)}}, pages 1426--1436.
  Association for Computational Linguistics.

\bibitem[{Lake et~al.(2017)Lake, Ullman, Tenenbaum, and
  Gershman}]{lake2017building}
Brenden~M. Lake, Tomer~D. Ullman, Joshua~B. Tenenbaum, and Samuel~J. Gershman.
  2017.
\newblock \href {https://doi.org/10.1017/S0140525X16001837} {Building machines
  that learn and think like people}.
\newblock \emph{Behavioral and Brain Sciences}, 40.

\bibitem[{Liu et~al.(2019{\natexlab{a}})Liu, Schwartz, and
  Smith}]{liu2019inoculation}
Nelson~F. Liu, Roy Schwartz, and Noah~A. Smith. 2019{\natexlab{a}}.
\newblock \href {https://www.aclweb.org/anthology/N19-1225} {Inoculation by
  fine-tuning: A method for analyzing challenge datasets}.
\newblock In \emph{Proceedings of the 2019 Conference of the North {A}merican
  Chapter of the Association for Computational Linguistics: Human Language
  Technologies, Volume 1 (Long and Short Papers)}, pages 2171--2179,
  Minneapolis, Minnesota. Association for Computational Linguistics.

\bibitem[{Liu et~al.(2019{\natexlab{b}})Liu, Ott, Goyal, Du, Joshi, Chen, Levy,
  Lewis, Zettlemoyer, and Stoyanov}]{liu2019roberta}
Yinhan Liu, Myle Ott, Naman Goyal, Jingfei Du, Mandar Joshi, Danqi Chen, Omer
  Levy, Mike Lewis, Luke Zettlemoyer, and Veselin Stoyanov. 2019{\natexlab{b}}.
\newblock \href {http://arxiv.org/abs/1907.11692} {{RoBERTa}: {A} robustly
  optimized {BERT} pretraining approach}.
\newblock arXiv preprint 1907.11692.

\bibitem[{MacWhinney(2000)}]{macwhinney2000tools}
Brian MacWhinney. 2000.
\newblock \emph{The {CHILDES} Project: Tools for Analyzing Talk. Third
  edition.}
\newblock Lawrence Erlbaum Associates, Mahwah, NJ.

\bibitem[{Marelli et~al.(2014)Marelli, Menini, Baroni, Bentivogli, Bernardi,
  and Zamparelli}]{marelli2014sick}
Marco Marelli, Stefano Menini, Marco Baroni, Luisa Bentivogli, Raffaella
  Bernardi, and Roberto Zamparelli. 2014.
\newblock \href
  {http://www.lrec-conf.org/proceedings/lrec2014/pdf/363_Paper.pdf} {A {SICK}
  cure for the evaluation of compositional distributional semantic models}.
\newblock In \emph{Proceedings of the Ninth International Conference on
  Language Resources and Evaluation ({LREC}'14)}, pages 216--223, Reykjavik,
  Iceland. European Language Resources Association (ELRA).

\bibitem[{Marr(1982)}]{marr1982vision}
David Marr. 1982.
\newblock \emph{Vision: A computational investigation into the human
  representation and processing of visual information}.
\newblock New York: Freeman.

\bibitem[{Marvin and Linzen(2018)}]{marvinlinzen18}
Rebecca Marvin and Tal Linzen. 2018.
\newblock \href {https://doi.org/10.18653/v1/D18-1151} {Targeted syntactic
  evaluation of language models}.
\newblock In \emph{Proceedings of the 2018 Conference on Empirical Methods in
  Natural Language Processing}, pages 1192--1202, Brussels, Belgium.
  Association for Computational Linguistics.

\bibitem[{McCoy et~al.(2019{\natexlab{a}})McCoy, Min, and
  Linzen}]{mccoy2019berts}
R.~Thomas McCoy, Junghyun Min, and Tal Linzen. 2019{\natexlab{a}}.
\newblock \href {http://arxiv.org/abs/1911.02969} {Berts of a feather do not
  generalize together: Large variability in generalization across models with
  similar test set performance}.

\bibitem[{McCoy et~al.(2019{\natexlab{b}})McCoy, Pavlick, and
  Linzen}]{mccoy2019right}
R.~Thomas McCoy, Ellie Pavlick, and Tal Linzen. 2019{\natexlab{b}}.
\newblock \href {https://doi.org/10.18653/v1/P19-1334} {Right for the wrong
  reasons: Diagnosing syntactic heuristics in natural language inference}.
\newblock In \emph{Proceedings of the 57th Annual Meeting of the Association
  for Computational Linguistics}, pages 3428--3448, Florence, Italy.
  Association for Computational Linguistics.

\bibitem[{Min et~al.(2020)Min, McCoy, Das, Pitler, and
  Linzen}]{min2020augmentation}
Junghyun Min, R.~Thomas McCoy, Dipanjan Das, Emily Pitler, and Tal Linzen.
  2020.
\newblock \href {https://arxiv.org/abs/2004.11999} {Syntactic data augmentation
  increases robustness to inference heuristics}.
\newblock In \emph{Proceedings of the 58th Annual Meeting of the Association
  for Computational Linguistics}, Seattle, Washington. Association for
  Computational Linguistics.

\bibitem[{Nangia and Bowman(2019)}]{nangia2019bowman}
Nikita Nangia and Samuel~R. Bowman. 2019.
\newblock \href {https://doi.org/10.18653/v1/P19-1449} {Human vs. muppet: A
  conservative estimate of human performance on the {GLUE} benchmark}.
\newblock In \emph{Proceedings of the 57th Annual Meeting of the Association
  for Computational Linguistics}, pages 4566--4575, Florence, Italy.
  Association for Computational Linguistics.

\bibitem[{Nie et~al.(2019)Nie, Williams, Dinan, Bansal, Weston, and
  Kiela}]{nie2019adversarial}
Yixin Nie, Adina Williams, Emily Dinan, Mohit Bansal, Jason Weston, and Douwe
  Kiela. 2019.
\newblock \href {https://arxiv.org/abs/1910.14599} {Adversarial {NLI}: A new
  benchmark for natural language understanding}.
\newblock arXiv preprint 1910.14599.

\bibitem[{Peters et~al.(2018)Peters, Neumann, Iyyer, Gardner, Clark, Lee, and
  Zettlemoyer}]{peters2018elmo}
Matthew Peters, Mark Neumann, Mohit Iyyer, Matt Gardner, Christopher Clark,
  Kenton Lee, and Luke Zettlemoyer. 2018.
\newblock \href {https://www.aclweb.org/anthology/N18-1202/} {Deep
  contextualized word representations}.
\newblock In \emph{Proceedings of the 2018 Conference of the North American
  Chapter of the Association for Computational Linguistics: Human Language
  Technologies, Volume 1 (Long Papers)}, pages 2227--2237. Association for
  Computational Linguistics.

\bibitem[{Poliak et~al.(2018)Poliak, Naradowsky, Haldar, Rudinger, and
  Van~Durme}]{poliak2018hypothesis}
Adam Poliak, Jason Naradowsky, Aparajita Haldar, Rachel Rudinger, and Benjamin
  Van~Durme. 2018.
\newblock \href {https://www.aclweb.org/anthology/S18-2023/} {Hypothesis only
  baselines in natural language inference}.
\newblock In \emph{{Proceedings of the Seventh Joint Conference on Lexical and
  Computational Semantics}}, pages 180--191. Association for Computational
  Linguistics.

\bibitem[{Raffel et~al.(2019)Raffel, Shazeer, Roberts, Lee, Narang, Matena,
  Zhou, Li, and Liu}]{raffel2019exploring}
Colin Raffel, Noam Shazeer, Adam Roberts, Katherine Lee, Sharan Narang, Michael
  Matena, Yanqi Zhou, Wei Li, and Peter~J Liu. 2019.
\newblock \href {https://arxiv.org/abs/1910.10683} {Exploring the limits of
  transfer learning with a unified text-to-text transformer}.
\newblock arXiv preprint 1910.10683.

\bibitem[{Roy et~al.(2015)Roy, Frank, DeCamp, Miller, and
  Roy}]{roy2015predicting}
Brandon~C. Roy, Michael~C. Frank, Philip DeCamp, Matthew Miller, and Deb Roy.
  2015.
\newblock \href {https://doi.org/10.1073/pnas.1419773112} {Predicting the birth
  of a spoken word}.
\newblock \emph{Proceedings of the National Academy of Sciences},
  112(41):12663--12668.

\bibitem[{van Schijndel et~al.(2019)van Schijndel, Mueller, and
  Linzen}]{vanschijndel2019quantity}
Marten van Schijndel, Aaron Mueller, and Tal Linzen. 2019.
\newblock \href {https://www.aclweb.org/anthology/D19-1592/} {Quantity
  doesn{'}t buy quality syntax with neural language models}.
\newblock In \emph{Proceedings of the 2019 Conference on Empirical Methods in
  Natural Language Processing and the 9th International Joint Conference on
  Natural Language Processing (EMNLP-IJCNLP)}, pages 5831--5837, Hong Kong,
  China. Association for Computational Linguistics.

\bibitem[{Schwartz et~al.(2019)Schwartz, Dodge, and Smith}]{schwartz2019green}
Roy Schwartz, Jesse Dodge, and Noah~A. Smith. 2019.
\newblock \href {https://arxiv.org/abs/1907.10597} {Green {AI}}.
\newblock arXiv preprint 1907.10597.

\bibitem[{Sprouse et~al.(2013)Sprouse, Sch{\"u}tze, and
  Almeida}]{sprouse2013comparison}
Jon Sprouse, Carson~T Sch{\"u}tze, and Diogo Almeida. 2013.
\newblock \href {https://doi.org/10.1016/j.lingua.2013.07.002} {A comparison of
  informal and formal acceptability judgments using a random sample from
  {Linguistic Inquiry} 2001--2010}.
\newblock \emph{Lingua}, 134:219--248.

\bibitem[{Suhr et~al.(2019)Suhr, Zhou, Zhang, Zhang, Bai, and
  Artzi}]{suhr2019nlvr2}
Alane Suhr, Stephanie Zhou, Ally Zhang, Iris Zhang, Huajun Bai, and Yoav Artzi.
  2019.
\newblock \href {https://www.aclweb.org/anthology/P19-1644} {A corpus for
  reasoning about natural language grounded in photographs}.
\newblock In \emph{Proceedings of the Annual Meeting of the Association for
  Computational Linguistics}, pages 6418--6428, Florence, Italy. Association
  for Computational Linguistics.

\bibitem[{Vaswani et~al.(2017)Vaswani, Shazeer, Parmar, Uszkoreit, Jones,
  Gomez, Kaiser, and Polosukhin}]{vaswani2017attention}
Ashish Vaswani, Noam Shazeer, Niki Parmar, Jakob Uszkoreit, Llion Jones,
  Aidan~N Gomez, \L~ukasz Kaiser, and Illia Polosukhin. 2017.
\newblock \href
  {http://papers.nips.cc/paper/7181-attention-is-all-you-need.pdf} {Attention
  is all you need}.
\newblock In I.~Guyon, U.~V. Luxburg, S.~Bengio, H.~Wallach, R.~Fergus,
  S.~Vishwanathan, and R.~Garnett, editors, \emph{{Advances in Neural
  Information Processing Systems 30}}, pages 5998--6008. Curran Associates,
  Inc.

\bibitem[{Wang et~al.(2019{\natexlab{a}})Wang, Pruksachatkun, Nangia, Singh,
  Michael, Hill, Levy, and Bowman}]{superglue}
Alex Wang, Yada Pruksachatkun, Nikita Nangia, Amanpreet Singh, Julian Michael,
  Felix Hill, Omer Levy, and Samuel~R. Bowman. 2019{\natexlab{a}}.
\newblock \href {http://arxiv.org/abs/1905.00537} {{SuperGLUE}: {A} stickier
  benchmark for general-purpose language understanding systems}.
\newblock arXiv preprint 1905.00537.

\bibitem[{Wang et~al.(2019{\natexlab{b}})Wang, Singh, Michael, Hill, Levy, and
  Bowman}]{wang2019glue}
Alex Wang, Amanpreet Singh, Julian Michael, Felix Hill, Omer Levy, and
  Samuel~R. Bowman. 2019{\natexlab{b}}.
\newblock \href {https://openreview.net/forum?id=rJ4km2R5t7} {{GLUE:} {A}
  multi-task benchmark and analysis platform for natural language
  understanding}.
\newblock In \emph{7th International Conference on Learning Representations,
  {ICLR} 2019, New Orleans, LA, USA, May 6-9, 2019}.

\bibitem[{Warstadt et~al.(2019)Warstadt, Parrish, Liu, Mohananey, Peng, Wang,
  and Bowman}]{warstadt2019blimp}
Alex Warstadt, Alicia Parrish, Haokun Liu, Anhad Mohananey, Wei Peng, Sheng-Fu
  Wang, and Samuel~R. Bowman. 2019.
\newblock \href {https://arxiv.org/abs/1912.00582} {{BLiMP}: A benchmark of
  linguistic minimal pairs for {E}nglish}.
\newblock arXiv preprint 1912.00582.

\bibitem[{Wilcox et~al.(2019)Wilcox, Qian, Futrell, Ballesteros, and
  Levy}]{wilcox2019structural}
Ethan Wilcox, Peng Qian, Richard Futrell, Miguel Ballesteros, and Roger Levy.
  2019.
\newblock \href {https://www.aclweb.org/anthology/N19-1334/} {Structural
  supervision improves learning of non-local grammatical dependencies}.
\newblock In \emph{Proceedings of the 2019 Conference of the North {A}merican
  Chapter of the Association for Computational Linguistics: Human Language
  Technologies, Volume 1 (Long and Short Papers)}, pages 3302--3312,
  Minneapolis, Minnesota. Association for Computational Linguistics.

\bibitem[{Williams et~al.(2018)Williams, Nangia, and
  Bowman}]{williams2018multinli}
Adina Williams, Nikita Nangia, and Samuel Bowman. 2018.
\newblock \href {http://aclweb.org/anthology/N18-1101} {A broad-coverage
  challenge corpus for sentence understanding through inference}.
\newblock In \emph{Proceedings of the 2018 Conference of the North American
  Chapter of the Association for Computational Linguistics: Human Language
  Technologies, Volume 1 (Long Papers)}, pages 1112--1122. Association for
  Computational Linguistics.

\bibitem[{Yang et~al.()Yang, Dai, Yang, Carbonell, Salakhutdinov, and
  Le}]{yang2019xlnet}
Zhilin Yang, Zihang Dai, Yiming Yang, Jaime~G. Carbonell, Ruslan Salakhutdinov,
  and Quoc~V. Le.
\newblock \href {https://arxiv.org/abs/1906.08237} {{XLNet}: Generalized
  autoregressive pretraining for language understanding}.
\newblock arXiv preprint 1906.08237.

\bibitem[{Yogatama et~al.(2019)Yogatama, d'Autume, Connor, Kocisky,
  Chrzanowski, Kong, Lazaridou, Ling, Yu, Dyer et~al.}]{yogatama2019learning}
Dani Yogatama, Cyprien de~Masson d'Autume, Jerome Connor, Tomas Kocisky, Mike
  Chrzanowski, Lingpeng Kong, Angeliki Lazaridou, Wang Ling, Lei Yu, Chris
  Dyer, et~al. 2019.
\newblock \href {https://arxiv.org/abs/1901.11373} {Learning and evaluating
  general linguistic intelligence}.
\newblock arXiv preprint 1901.11373.

\bibitem[{Zellers et~al.(2018)Zellers, Bisk, Schwartz, and
  Choi}]{zellers2018swag}
Rowan Zellers, Yonatan Bisk, Roy Schwartz, and Yejin Choi. 2018.
\newblock \href {https://doi.org/10.18653/v1/D18-1009} {{SWAG}: A large-scale
  adversarial dataset for grounded commonsense inference}.
\newblock In \emph{Proceedings of the 2018 Conference on Empirical Methods in
  Natural Language Processing}, pages 93--104, Brussels, Belgium. Association
  for Computational Linguistics.

\end{thebibliography}
\bibliographystyle{acl_natbib}

\end{document}